\documentclass[10pt,twocolumn,letterpaper]{article}

\usepackage{wacv}
\usepackage{times}
\usepackage{epsfig}
\usepackage{graphicx}
\usepackage{amsmath}
\usepackage{amssymb}
\usepackage{makecell}
\usepackage{booktabs}

\wacvfinalcopy 

\def\wacvPaperID{****} 

\ifwacvfinal\fi
\begin{document}

\author{
	Shichao Xu\\
	Northwestern University\\
	{\tt\small ShichaoXu2023@u.northwestern.edu}
	\and
	Shuyue Lan\\
	Northwestern University\\
	{\tt\small shuyuelan2018@u.northwestern.edu}
	\and
	Qi Zhu\\
	Northwestern University\\
	{\tt\small qzhu@northwestern.edu}
}

\title{MaskPlus: Improving Mask Generation for Instance Segmentation}

\maketitle
\ifwacvfinal\thispagestyle{empty}\fi

\begin{abstract}
	Instance segmentation is a promising yet challenging topic in computer vision. Recent approaches such as Mask R-CNN typically divide this problem into two parts -- a detection component and a mask generation branch, and mostly focus on the improvement of the detection part. 
	In this paper, we present an approach that extends Mask R-CNN with five novel techniques for improving the mask generation branch and reducing the conflicts between the mask branch and the detection component in training. 
	These five techniques are independent to each other and can be flexibly utilized in building various instance segmentation architectures for increasing the overall accuracy. We demonstrate the effectiveness of our approach with tests on the COCO dataset.
\end{abstract}

\section{Introduction}
Instance segmentation is a promising and challenging task in vision, with potential applications in medical imaging~\cite{chen2017dcan, chen2016iterative}, autonomous vehicles~\cite{alhaija2017augmented, van2016instance}, smart city~\cite{sajjad2016leukocytes}, robotics~\cite{shvets2018automatic, bewley2014online}, etc. The problem has received significant interests in recent years. 
It can be viewed as a more complex case than semantic segmentation, as we not only need to segment and classify the objects, but also should identify each individual instance. 

In the literature, researchers have proposed a number of approaches for instance segmentation. 
One popular idea is to cluster the similar contents in the image. For instance, Bert et al.~\cite{de2017semantic} explored an approach of using convolutional neural networks (CNNs) to produce a representation that can be easily clustered into instances. Alireza et al.~\cite{fathi2017semantic} proposed a fully convolutional embedding model to segment the instances by computing the likelihood of two pixels belonging to same object. 
Alejandro et al.'s work in~\cite{newell2017associative} also received significant interests. It introduced a new algorithm named associative embedding, which can teach networks to output joint detection and group assignments in a single stage. 
Similar idea also appeared in~\cite{silberman2014instance, kong2018recurrent, zhang2015monocular, uhrig2016pixel}.

One of the most successful object detection methods is Faster R-CNN~\cite{ren2015faster}, which extended the work in fast R-CNN~\cite{girshick2015fast} by adding the region proposal network (RPN) to speed up the region proposal process. This approach has also been applied to instance segmentation and led to a number of proposal-based methods. For instance, Dai et al.~\cite{dai2016instance} fused the detection and classification steps with a cascade network, and obtained the top result in the 2015 MS-COCO instance segmentation challenge. Xu et al.~\cite{xu2017gland} extracted regional, location and boundary features from gland histology images and created a CNN to identify the object individuals. Later, Li et al.~\cite{li2017fully} adopted the idea from InstanceFCN~\cite{dai2016instance} and used position-sensitive score map to perform object segmentation and detection at the same time.

Recently, Mask R-CNN~\cite{he2017mask} and its extensions such as PANet~\cite{liu2018path} take advantage of the Faster R-CNN detection framework for the further improvement of instance segmentation. 
According to them, an instance segmentation task can be viewed as the combination of a detection problem and a segmentation problem. Their solutions first apply a detection component and then a mask generation branch, where segmentation is performed based on the Region of Interest (RoI) features.
Such approaches avoid the problem of spurious edges in the FCIS method~\cite{li2017fully}, and also eliminate the misalignment in the RoI-pooling process and gets exact spatial locations. 
They represent the state-of-the-art results on instance segmentation.

However, most of these works focus on improving the detection component, while there are still \emph{significant limitations in the mask branch that require further improvements}: 
1) the mask branch can only get the features from the RoI and may suffer from the loss of global semantic information; 2) imperfect bounding boxes degrade the overall performance in the mask branch; 3) simple mask branch architecture with one deconvolutional layer and the lack of boundary refinement lead to coarse results; and 4) conflicts in multitask training (due to different learning pace of each part) may cause performance degradation.

To address these limitations, we developed a new instance segmentation framework \textbf{MaskPlus} in this work, which extends Mask R-CNN with five novel techniques to boost the performance of mask generation: 1) contextual fusion, 2) deconvolutional pyramid module, 3) boundary refinement, 4) quasi-multitask learning, and 5) biased training. More specifically, the contributions of this work include:
\begin{itemize}
	\item We created novel techniques -- contextual fusion, quasi-multitask learning and biased training to incorporate global information into the mask generation branch, better supervised mask training and reduce the conflicts that happen in multitask training.
	\item We further extend the existing techniques, including boundary refinement, deconvolutional pyramid module, to improve the overall performance and get finer mask results.
	\item We tested our approach on the COCO instance segmentation dataset \cite{lin2014microsoft} and show our competitive results on CodaLab leaderboard. We conducted ablation studies to illustrate the efficacy of each technique, and also evaluated the overall instance segmentation performance. It is demonstrated that our MaskPlus is effective and can achieve the state-of-the-art results.
\end{itemize}


The rest of the paper is organized as follows. Section~\ref{sec:related_work} introduces related works in more details. Section~\ref{sec:framework} presents our proposed MaskPlus framework, with details for each of the five techniques. Section~\ref{sec:experiments} shows the experimental results of our ablation studies and overall evaluation. Section~\ref{sec:conclusion} concludes the paper.

\section{Related Work}
\label{sec:related_work}

\subsection{Instance Segmentation}
One common approach for instance segmentation is clustering, which gathers the similar pixels to form the instances. Liang et al.~\cite{liang2015proposal} used proposal free network to generate the coordinates of the instance bounding box and the confidence scores of different categories for each pixel, and then added clustering as the post-processing module to generate the instance results. Bert et al.~\cite{de2017semantic} presented an approach to produce a representation from CNNs that can be easily clustered into instances by applying the mean-shift algorithm to obtain cluster centers. Alireza et al.~\cite{fathi2017semantic} learned a similarity metric by creating a deep embedding model and grouped similar pixels together. Similar ideas can also be found in~\cite{newell2017associative, silberman2014instance, kong2018recurrent, zhang2015monocular, uhrig2016pixel}.

Another type of approach leverages the success from object detection methods such as the Faster R-CNN model~\cite{ren2015faster} and its region proposal network. Dai et al.~\cite{dai2016instance} won 2015 MS-COCO instance segmentation challenge by building a cascade network and connected the steps of detection and segmentation. Xu et al.~\cite{xu2017gland} split the entire instance segmentation task into sub-tasks and generate regional, location and boundary features from gland histology images to classify the objects. Li et al.~\cite{li2017fully} applied RoIs onto the position-sensitive score map to address instance segmentation. 
Mask R-CNN~\cite{he2017mask} presented one of the most promising methods in recent years. It is built on the Faster R-CNN framework and adds an FCN backend after the RoIs as the mask generation branch. It also solves the problem of misalignment in the RoI-pooling process with a RoIAlign layer. Later, Liu et al.~\cite{liu2018path} extended the Mask R-CNN model by improving the backbone networks, adding bottom-up path augmentation and adding fully-connected paths from the features of RoIs to the features after deconvolutional layer. These extensions mostly focus on the detection component and do not address the limitations in the mask generation branch, which is the focus of our paper.

\subsection{Semantic Segmentation}

For the related problem of semantic segmentation, deep learning methods have been widely used. 
In Long et al.'s FCN model~\cite{long2015fully}, end-to-end algorithm was introduced and deconvolution was utilized for up-sampling. 
Later, Badrinarayanan et al.~\cite{badrinarayanan2015segnet} improved the method by recording the position information during pooling and applied it in the up-sampling process. U-Net~\cite{ronneberger2015u} enhanced the information delivery from earlier layers to the higher layers with specially designed U-shape network framework. Yu et al.~\cite{yu2015multi} applied the dilated convolutions for semantic segmentation. Chen et al.~\cite{chen2018deeplab} adopted the similar idea, and used atrous spatial pyramid pooling (ASPP) and fully connected CRF to make model more accurate. Other works include RefineNet~\cite{lin2017refinenet}, PSPNet~\cite{zhao2017pyramid}, Large Kernel Matter~\cite{peng2017large}, DeepLab v3~\cite{chen2017rethinking}, etc.

\section{Proposed MaskPlus Framework}
\label{sec:framework}

In this section, we introduce the proposed MaskPlus Framework, and explain the details of the five techniques and how they address the limitations of previous mask generation methods in instance segmentation.


\begin{figure*}[h]
	\begin{center}
		\includegraphics[width=0.95\linewidth]{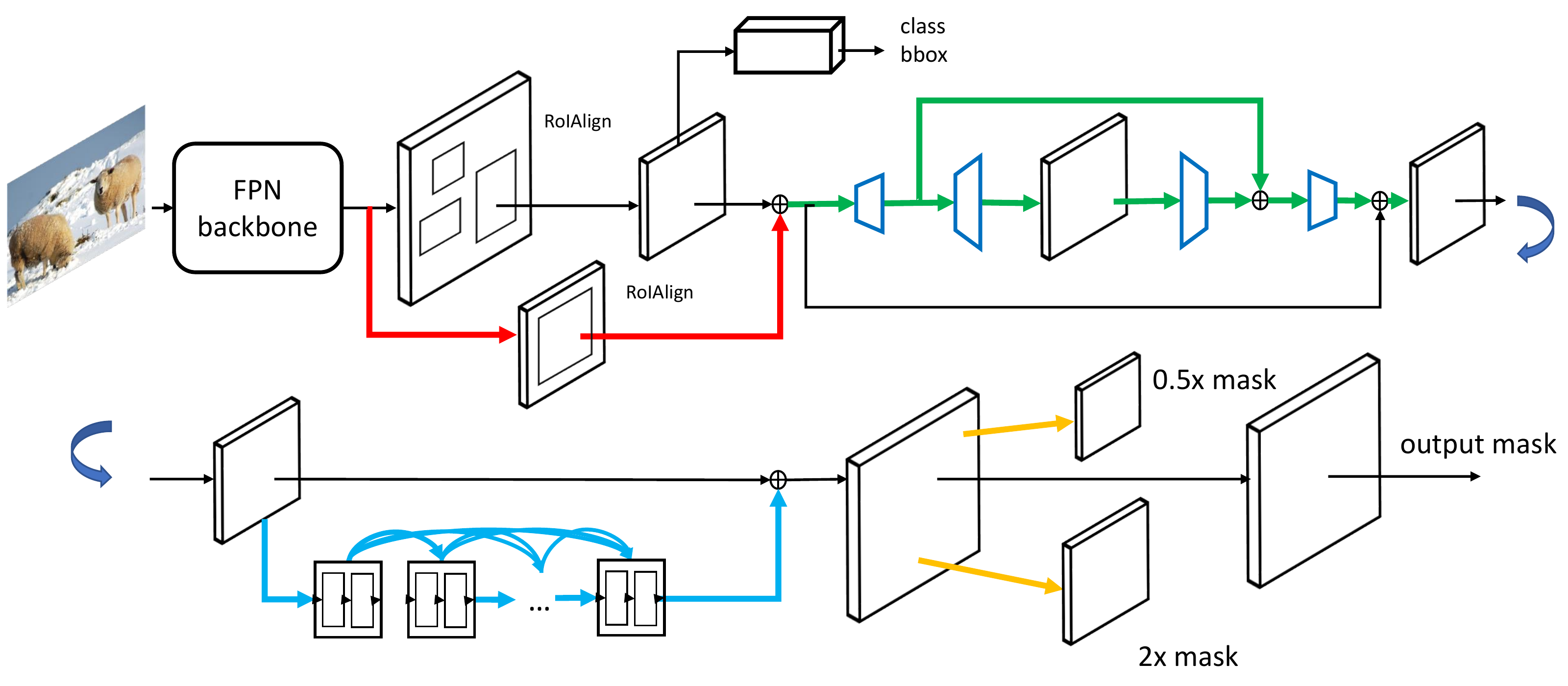}
	\end{center}
	\caption{\textbf{Overview of the MaskPlus Framework.} Given an input image, MaskPlus outputs a generated segmentation mask. It extends the Mask R-CNN framework with various techniques on mask generation, including 
		contextual fusion (in red), deconvolutional pyramid module (in green), improved boundary refinement (in blue), quasi-multitask learning (in yellow), and biased training (not shown in the figure). The figure is best viewed in color. }
	\label{framework}
\end{figure*}

Figure~\ref{framework} shows an overview of MaskPlus, which extends the Mask R-CNN framework. 
First, a Faster R-CNN model with FPN structure is applied as the backbone. It has a branch with two detection related outputs –- a classification output and a bounding box output. The RoIAlign technique is used to replace the original RoI-pooling process to give the pixel-to-pixel alignment for the results. Then the mask generation branch is applied to the output of RoI features to generate the mask output. The novelty of MaskPlus resides on the five techniques for improving the mask generation branch, which are introduced below.

\subsection{Contextual Fusion}

\begin{figure}[h]
	\centering
	\vspace{-6pt}
	\includegraphics[width=1.0\linewidth]{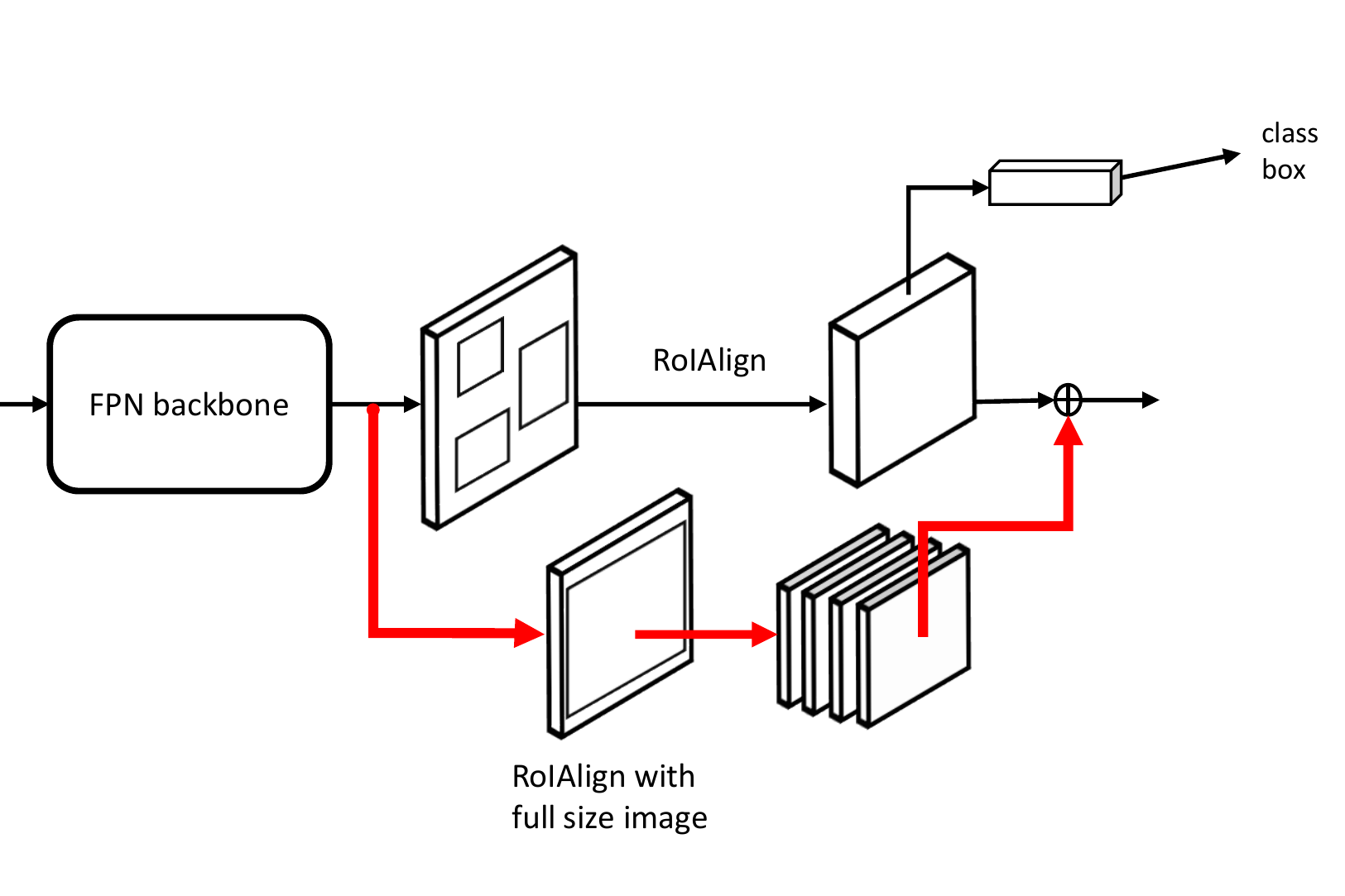}
	\caption{Contextual Fusion.}
	\label{Contextual}
\end{figure}

In the mask branch of Mask R-CNN, the features are only generated from the RoIs. We believe that this could limit the generation of mask prediction because of the lacking of contextual information: First, the semantic information of one object could also contain other objects, as they may have spatial relation, semantic relation, etc. For instance, a car may exist on a road but not in the sky (not yet), and thus the road could have some part of the information of car objects. It may not be effective to separate the objects and segment them one by one. Second, the RoIs are not always perfect. They may not contain some parts of the target object, which could make it difficult to segment these fragmentary objects. Third, some parts of the object that do not belong to the RoI's category may exist at the boundary. They may be mis-classified and disturb the mask generation, as they also lack their own semantic meaning and the neural networks cannot recognize them. 

To address these challenges, we create a new branch from the features just before the RoIAlign module, as shown with the red lines in Figure~\ref{framework}. 
In the fusion procedure of global features, we take FPN last-layer features and apply only one full-image-size proposal at a newly-created RoIAlign layer. Then through 3 conv layers (kernel size is 3, stride is 1, filter number are 512, 256, 256, respectively), the outcomes on this new branch (the red-line path in Figure~\ref{framework}) will be added to the RoIAlign features from the original mask branch. The newly created RoIAlign layer and the old one have the same configuration and size of feature outputs.
Such fusion helps the RoI features to get more contextual information.

Note that our approach is very different from the Fully-connected Fusion technique in ~\cite{liu2018path}.
In ~\cite{liu2018path}, the input features forwarded to the up-sampling layer contain two concatenated parts -- the output features from the ROI Align layer ($f1$) and the output features of the Fully-connected Fusion layer ($f2$). However, the input of the Fully-connected Fusion layer is just $f1$, which is a set of features of ROIs. It has lost the global spatial relationship between these ROIs, and the spatial information it contained is limited within individual proposals. In our approach, the input features are full-size features before the ROI Align layer, which contain the global spatial relationship of each object. Besides, the motivation and used methods are also different in the two approaches. \cite{liu2018path} aims to utilize the strong points of fully connected layer, which does not exist in our module.
\subsection{Deconvolutional Pyramid Module}

\begin{figure}[h]
	\centering
	\vspace{-6pt}
	\includegraphics[width=1.0\linewidth]{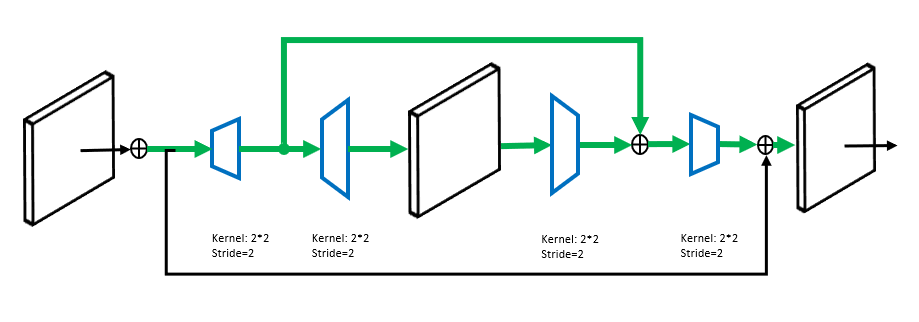}
	\caption{Deconvolutional pyramid module.}
	\label{pyramid}
\end{figure}

Motivated by the structure of Feature Pyramid Networks(FPN)~\cite{lin2017feature}, which builds a pyramid module to fuse multi-level features in the early stages of the network, we define a deconvolutional pyramid module as a set of deconvolutional layers (stride = 2) followed by the equal numbers of convolutional layers (stride = 2), as shown with the green lines in Figure~\ref{framework}. Our design is different from the FPN though -- instead of first applying down-sampling and then up-sampling, our module up-samples first and then down-samples, as shown with more details in Figure~\ref{pyramid}. 

We believe that, instead of just delivering the original features to a single up-sampling layer, our module can finetune existing features and combine multi-level semantic meanings together to generate better mask prediction.
We observe that adding this module can improve the mask accuracy among all sizes (S, M, L) in our experiments (details in Section~\ref{sec:experiments}). 

\subsection{Improved Boundary Refinement}

In the mask generation of instance segmentation, we often observe blurring boundaries --
as the larger scores in the feature map mainly focus on the center part of the objects’ score map rather than staying at the boundary, it is often difficult to clearly identify the mask boundary. 
To address this challenge, we propose to learn the boundary by adding another branch, as shown with the blue lines in Figure~\ref{framework} and detailed in Figure~\ref{boundary}.
\begin{figure}
	\centering
	\includegraphics[width=\linewidth]{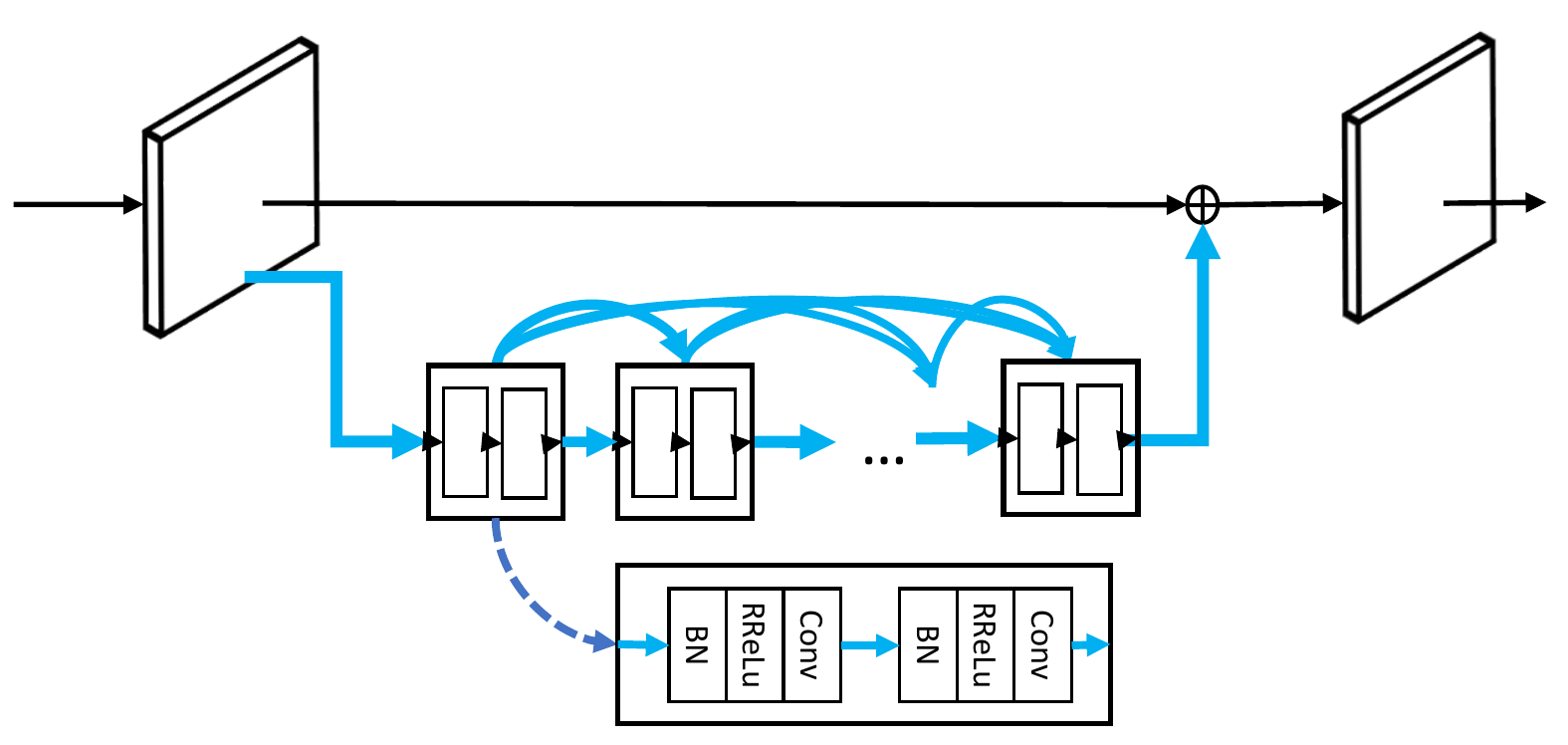}
	\caption{Improved boundary refinement.}
	\label{boundary}
\end{figure}

Our approach is inspired by the work from Peng et al.~\cite{peng2017large}, which uses a residual block to sharpen the boundary. The difference is that we think it is insufficient to just learn the boundary with only two convolutional layers that act as a single residual block. Instead, we create a branch that consists of several convolutional modules with dense connections for better learning ability to refine the boundary information.
In our experiments, we demonstrate the effectiveness of this improved boundary refinement by comparing its segmentation performance with both the original boundary refinement in~\cite{peng2017large} and the model without boundary refinement. 

\subsection{Quasi-multitask Learning}

Deep convolutional neural networks are thought to be strongly rotation invariant and scale invariant, however using them may still be insufficient to provide the desired robustness. 
Thus, researchers have proposed image-augmentation techniques (that include pre-processing for image rotations and rescaling to multiple scales) and feature rescaling and extracting methods such as FPN~\cite{lin2017feature}.   

In this work, instead of learning the features from different scales of images (augmentation at the top of the network) or creating a feature pyramid architecture (augmentation in the middle of a network), we consider using the labels in different scales as a guide for helping the networks to enhance scale invariant ability (augmentation at the end of the network).
We develop a quasi-multitask learning approach, aka quasi-multitask as we perform similar tasks, to increase the robustness of our framework, as shown with the yellow lines in Figure~\ref{framework}. 

More specifically, we tested the effect of combined training on the original size of mask (resized to 28 * 28), with the 0.5x size of mask (14 * 14), or the 2x size of mask(56 * 56). These branches are parallel to each other and inserted after ROIAlign features (See Section 4.7 for more details). And the most important thing is that -- regardless of what other scales we add, we never use these different scale branches as the output of the final mask. That is to say, the parameters or FLOPs never increase in test stage and the output scale is never changed, while the performance increases. To the best of our knowledge, we are the first to explore this approach.

\subsection{Biased Training}

The original training strategy in Mask R-CNN trains the detection component and the mask generation branch together. In the Faster R-CNN's architecture, the RPN could be trained in advance, however it still does not solve the problem that the mask branch may not get a good RoI feature and thus provide ineffective feedback to the early stages of the training process. In some cases, the feedback from the mask branch may even disturb the training of the detection component, and then the disturbed detection results will have a negative impact on mask branch itself in the later stages.

In this work, we try to lower the influence of the training of the mask branch in the early stages, while still keep the end-to-end learning pattern.
First, we multiply the loss in the mask branch with a weight greater than 1, and define the multitask loss on each sampled RoI as $L = L_{cls} + L_{box} + \alpha(L_{mask})$. $L_{cls}$ and $L_{box}$ are the classification loss and bounding-box loss, respectively, as defined in~\cite{girshick2015fast}. $L_{mask}$ is the mask loss as defined in~\cite{he2017mask}. The parameter $\alpha$ is initially larger than $1$ (e.g., chosen as $1.5$ in our experiments). 

Intuitively, increasing one part of the loss seems to address it and makes it better. But in our design, it happens in opposite direction. The novelty that should be addressed is that using such loss function has the same effect as increasing the learning rate of the mask branch, which will force the mask branch to converge faster in the early stages. In this way, the potential negative influence of the mask branch at the early stages can be largely reduced. Then, $\alpha$ is set to $1$ during the normal training process in the later stages. This is because this technique also cause worse mask result in the long run. Thus we need another stage of normal training to mitigate.


\section{Experimental Results}
\label{sec:experiments}

\subsection{Dataset}

All of our experiments are performed on the challenging COCO dataset~\cite{lin2014microsoft}, which is also used by Mask R-CNN. The dataset has 115k training images and 5k validation images on 80 object categories. It also contains 41k test images for online testing, whose ground-truth labels are not publicly available. Our framework is trained on the train-2017 subset and perform the ablation study on the val-2017 subset. The standard COCO metrics includes $AP$ (averaged over IoU thresholds), $AP_{50}$, $AP_{75}$,  and $AP_S$, $AP_M$, $AP_L$ (AP for images at different scales: small, medium, large). The following experiments are evaluated using mask IoU, unless specifically mentioned as detection results ($AP^{bb}$).

\subsection{Training Configuration}

Our implementation is based on the Tensorpack framework~\cite{wu2016tensorpack}. We used the re-implemented version of Mask R-CNN in Tensorpack as the baseline, which shows better mask $AP$ than the original paper. The pretrained model is publicly available from the Tensorpack model zoo. Image centric training~\cite{girshick2015fast} is applied so that the images are resized to 800 pixels on the shorter edge, 1333 on the longer edge, without changing the aspect ratio. Each image has 512 sampled RoIs, and their positive to negatives ratio is 1:3. We use 8 Titan RTX GPUs in training and single image per GPU for 360000 iterations. The learning rate is 0.02, weight decay is 0.0001, and momentum is 0.9. Other configurations are the same as Mask R-CNN. The RPN is trained separately and do not share the weights with Mask R-CNN. In addition, all ablation studies are tested based on the ResNet-50-FPN backbone for faster training/testing speed.

\begin{figure*}[!h]
	\centering
	\setlength{\abovecaptionskip}{0pt}
	\setlength{\belowcaptionskip}{-16pt}
	\includegraphics[width=1.0\linewidth]{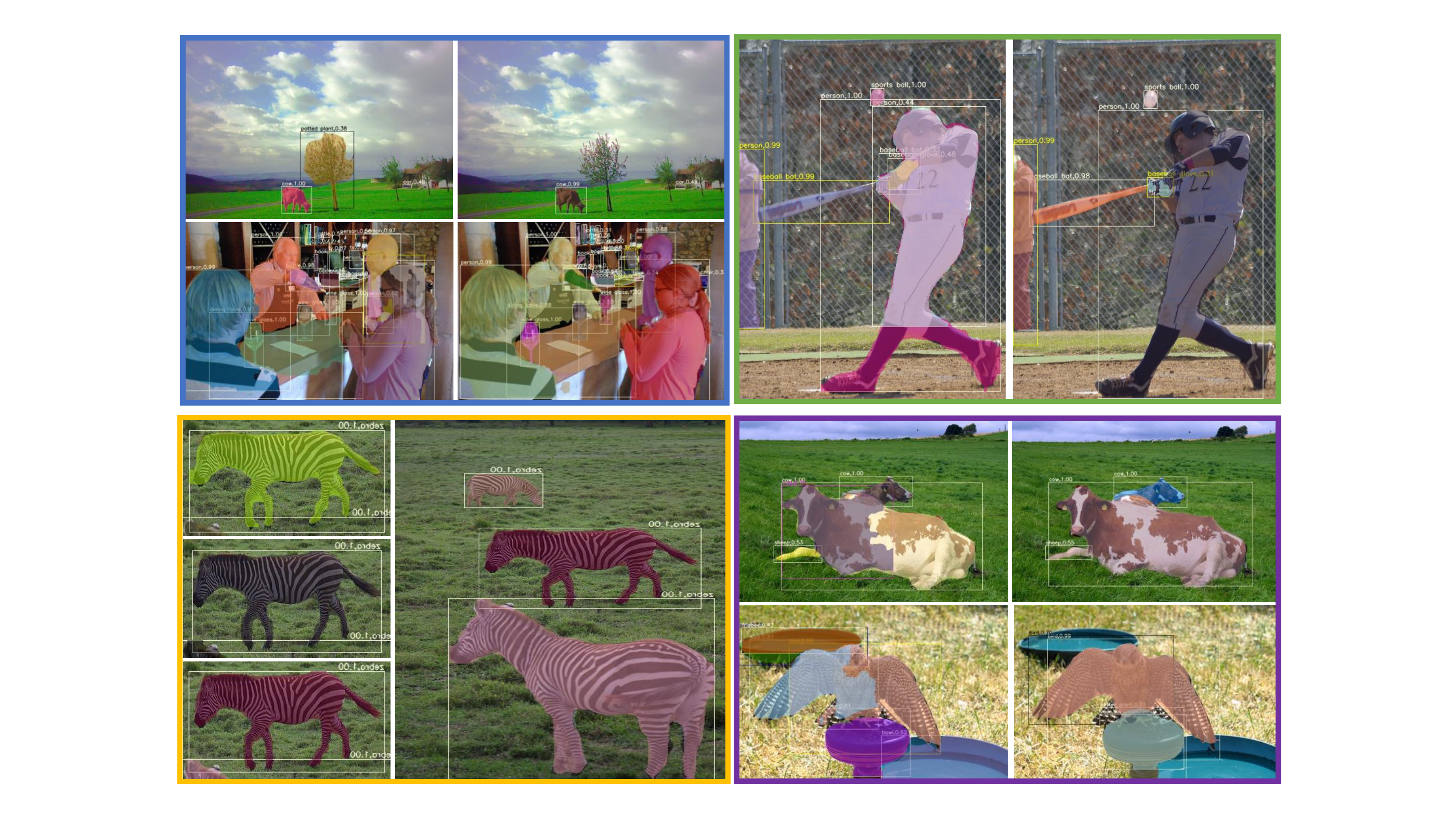}
	\caption{\small{1). In the blue box, Mask R-CNN is at left. MaskPlus with only contextual fusion is at right -- things like potted plant that should not appear in that scenario are removed, and the boundary fragment is correctly classified. 2). In the green box, Mask R-CNN is at left. MaskPlus with only deconvolutional pyramid module is at right -- redundancy is removed with the help of multi-scale feature information. 3). In the yellow box, Mask R-CNN is at upper left. Original boundary refinement is at middle left. Only using improved boundary refinement is at lower left. The right side is the result from using improved boundary refinement in MaskPlus -- the boundary is refined and has better quality from top to bottom. 4). In the purple box, Mask R-CNN is at left. MaskPlus with only quasi-multitask learning is at right -- multi-scale supervision prevents some misclassifications from single-scale supervision.}}
	\label{vis}
\end{figure*}

\subsection{Explanation with image visualization:}
Please zoom in to see the visual explanation for techniques about architectures in Figure~\ref{vis}. We can see the powerful effects that help Mask R-CNN do better work.

\subsection{Contextual Fusion}

Table~\ref{fusion} shows the results for adding the contextual fusion technique to the original Mask R-CNN framework, and we can see the improvements from using such technique. We observe that the configuration of a stack of convolutional layers between the new RoIAlign layer and the adding layer also affects the improvement. 
We tried different configurations and find the best to be [720, 512, 512, 256] (the numbers represent the filter numbers of these consecutive convolutional layers, number of layers change according to configuration), and even with deeper layers or wider filter numbers, the performance will decrease on the contrary (because simply adding more capacity on a big network will increase the difficulties on training optimization.).

With the help of the contextual fusion, the AP of mask branch increases from 35.1 to 35.5, while the precision of the detection component is not affected. More improvement is gained for middle- and large-size objects. This validates the function of the contextual fusion.

\begin{table}[!h]
    \footnotesize
	\centering
	\begin{tabular}{p{66pt}|p{15pt}|p{15pt}|p{15pt}|p{15pt}|p{15pt}|p{15pt}}
	    \hline
		Method & $AP$ & $AP_{50}$ & $AP_{75}$& $AP_{S}$ & $AP_{M}$ & $AP_{L}$\\
		\hline
		Mask R-CNN   & 35.1 & 56.6 & 37.5 & 18.4 & 38.4 & 48.3\\
		\hline
		{Mask R-CNN + Contextual fusion 256-256} & 35.3 & 56.7 & 37.6 & 18.6 & 38.6 & 48.6 \\
		\hline
		{Mask R-CNN + Contextual fusion 256-256-256} & 35.4 & 56.9 & 37.7 & 18.4 & 38.5 & 48.7 \\
		\hline
		{Mask R-CNN + Contextual fusion 720-512-512-256} & 35.5 & 57.0 & 37.8 & 18.5 & 38.7 & 48.8 \\
		\hline
	\end{tabular}
	\vspace{4pt}
	\caption{Ablation study of contextual fusion.}
	\label{fusion}
\end{table}

\subsection{Deconvolutional Pyramid Module}

In our deconvolutional pyramid module, two deconvolutional layers are followed by two convolutional layers, as shown in Figure~\ref{boundary}. They have strides of 2 and filter size of 256. The features are added instead of concatenated. The experimental results of applying this module are shown in Table~\ref{deconv}. We can see that using this technique increases the mask AP from 35.1 to 35.4. Moreover, the accuracy is mainly improved for small- and middle-size objects.

\begin{table}[!h]
	\footnotesize
	\centering
	\begin{tabular}{p{66pt}|p{15pt}|p{15pt}|p{15pt}|p{15pt}|p{15pt}|p{15pt}}
	    \hline
	    Method & $AP$ & $AP_{50}$ & $AP_{75}$& $AP_{S}$ & $AP_{M}$ & $AP_{L}$\\
		\hline
		{Mask R-CNN} & 35.1 & 56.6 & 37.5 & 18.4 & 38.4 & 48.3\\
		\hline
		{Mask R-CNN + Deconvolutional pyramid module} & 35.4 & 56.9 & 37.6 & 18.8 & 38.6 & 48.3 \\
		\hline
	\end{tabular}
	\vspace{4pt}
	\caption{Ablation study of deconvolutional pyramid module.}
	\label{deconv}
\end{table}

\subsection{Improved Boundary Refinement}

We choose a stack of convolutional modules with dense connections to learn the boundary deficiency. Specifically, each convolutional module consists of six layers in order: BatchNorm, PReLu, Conv (filters = 16), BatchNorm, PReLu, and Conv (filters = 4). The later modules will concatenate the input features from all previous modules as its own input features. The concatenation will be repeated for 4 modules. Figure~\ref{boundary} reflects this kind of design pattern.

We compare our improved boundary refinement method with the original Mask R-CNN and the Mask R-CNN with boundary refinement method described in~\cite{peng2017large}.
The results are presented in Table~\ref{boundary}. Our approach provides improvements on mask precision over both cases. Note that the precision is mostly improved for middle- and large-size objectives (no improvement for small-size objects). 

\begin{table}[!h]
	\footnotesize
	\centering
	\begin{tabular}{p{66pt}|p{15pt}|p{15pt}|p{15pt}|p{15pt}|p{15pt}|p{15pt}}
	\hline
		Method & $AP$ & $AP_{50}$ & $AP_{75}$& $AP_{S}$ & $AP_{M}$ & $AP_{L}$\\
		\hline
		{Mask R-CNN} & 35.1 & 56.6 & 37.5 & 18.4 & 38.4 & 48.3\\
		\hline
		{Mask R-CNN + Original boundary refinement} & 35.2 & 56.6 & 37.6 & 18.3 & 38.2 & 48.4 \\
		
		\hline
		{Mask R-CNN + Improved boundary refinement} & 35.5 & 56.9 & 38.0 & 18.3 & 38.8 & 49.1 \\
		\hline
	\end{tabular}
	\vspace{4pt}
	\caption{Ablation study of improved boundary refinement.}
	\label{boundary}
\end{table}

\subsection{Quasi-multitask Learning}

Instead of augmenting in front or middle of the networks, we developed the quasi-multitask learning technique to augment in the end (after the last layer of the networks). In the original Mask R-CNN configuration, the features from the RoIAlign layer will come across several convolutional layers with small filter numbers (these layers are the original settings in Mask R-CNN, noted to be $L$), and then they are up-sampled to 2x size (ground truth mask is resized to 28 * 28). Here we create another branch instead (parallel to $L$) follows the RoIAlign layer features. This branch has the same layers as $L$ except for one layer -- we will delete the last deconvolutional layer or add a deconvolutional layer at last with 2x upsampling scale to keep the output features to be 0.5x or 2x scale. Then the output mask will compute loss with 0.5x size of the original configuration (14 * 14) or 2x size (56 * 56). And for the most important thing is that the output mask is still the original branch, not the newly created one (whose purpose is only to help calculate the quasi-multitask learning loss). 
As shown in Table~\ref{multitask}, both the 0.5x and 2x quasi-multitask learning demonstrate improvements on the mask accuracy.

\begin{table}[!h]
	\footnotesize
	\centering
	\begin{tabular}{p{66pt}|p{15pt}|p{15pt}|p{15pt}|p{15pt}|p{15pt}|p{15pt}}
	\hline
		Method & $AP$ & $AP_{50}$ & $AP_{75}$& $AP_{S}$ & $AP_{M}$ & $AP_{L}$\\
		\hline
		{Mask R-CNN} & 35.1 & 56.6 & 37.5 & 18.4 & 38.4 & 48.3\\
		\hline
		{Mask R-CNN + 0.5x quasi-multitask learning} & 35.4 & 56.7 & 37.7 & 18.4 & 38.6 & 48.7 \\
		\hline
		{Mask R-CNN + 2x quasi-multitask learning} & 35.2 & 56.5 & 37.5 & 18.5 & 38.3 & 48.4 \\
		\hline
	\end{tabular}
	\vspace{4pt}
	\caption{Ablation study of quasi-multitask learning.}
	\label{multitask}
\end{table}

\subsection{Biased Training}

As introduced in Section~\ref{sec:framework}, we try to reduce the (negative) influence of the mask branch training in the first half stages on the detection component. Recall that the multitask loss is: $L = L_{cls} + L_{box} + \alpha(L_{mask})$,  and we initially set $\alpha = 1.5$ to increase the learning rate of the mask branch and force it to converge faster in early stages ($\alpha=1$ in the later stages). 
Table~\ref{biasdetection} shows the results of such biased training in detection, and Table~\ref{biasmask} shows its mask results. We can see that both detection component and mask branch gain benefits from biased training.

\begin{table}[!h]
	\footnotesize
	\centering
	\begin{tabular}{p{66pt}|p{15pt}|p{15pt}|p{15pt}|p{15pt}|p{15pt}|p{15pt}}
	\hline
		Method  & $AP^{bb}$ & $AP^{bb}_{50}$ & $AP^{bb}_{75}$& $AP^{bb}_{S}$ & $AP^{bb}_{M}$ & $AP^{bb}_{L}$\\
		\hline
		{Mask R-CNN} & 38.3 & 59.8 & 41.6 & 21.8 & 41.9 & 50.3\\
		\hline
		{Mask R-CNN + Biased training} & 38.6 & 60.0 & 41.8 & 21.7 & 42.1 & 51.3 \\
		\hline
	\end{tabular}
	\vspace{4pt}
	\caption{Ablation study of biased training on the detection component.}
	\label{biasdetection}
\end{table}
\begin{table}[!h]
	\footnotesize
	\centering
	\begin{tabular}{p{66pt}|p{15pt}|p{15pt}|p{15pt}|p{15pt}|p{15pt}|p{15pt}}
	\hline
		Method & $AP$ & $AP_{50}$ & $AP_{75}$& $AP_{S}$ & $AP_{M}$ & $AP_{L}$\\
		\hline
		{Mask R-CNN} & 35.1 & 56.6 & 37.5 & 18.4 & 38.4 & 48.3\\
		\hline
		{Mask R-CNN + Biased training} & 35.4 & 56.7 & 37.8 & 18.2 & 38.5 & 48.8 \\
		\hline
	\end{tabular}
	\vspace{4pt}
	\caption{Ablation study of biased training on the mask results.}
	\label{biasmask}
\end{table}
\begin{figure*}[!h]
	\begin{center}
		\includegraphics[width=0.9\linewidth]{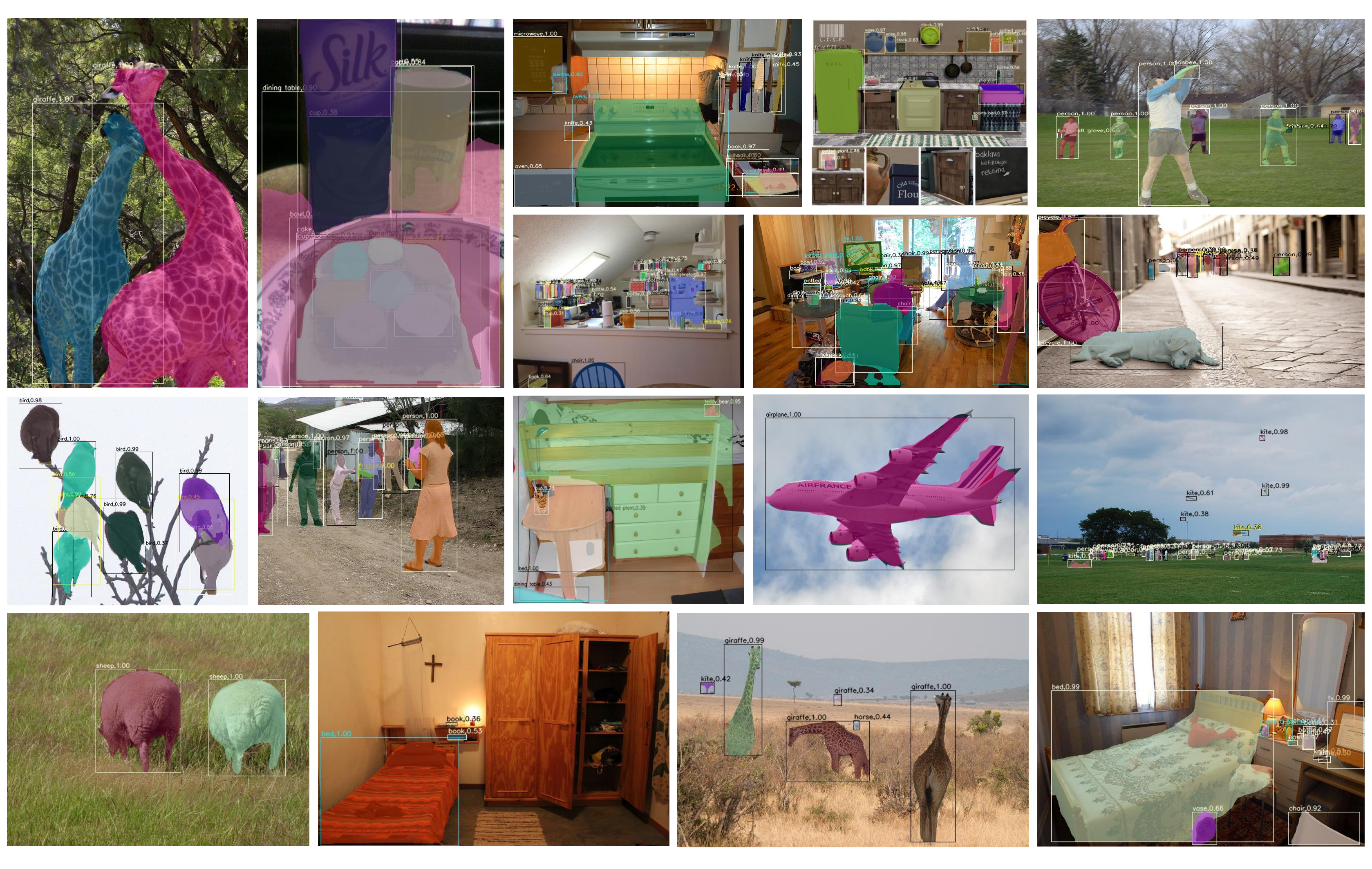}
	\end{center}
	\vspace{6pt}
	\caption{\textbf{MaskPlus visual results on the COCO dataset.} Masks are shown in color, and bounding box, category, and confidences are also presented. The figure is best viewed in color.}
	\label{example}
\end{figure*}
\begin{table*}[!h]

	\centering
	\small
	\begin{tabular}{l|l|l|l|l|l|l|l}
		Method & backbone & AP & $AP_{50}$ & $AP_{75}$& $AP_{S}$ & $AP_{M}$ & $AP_{L}$\\
		\hline
		\makecell[bl]{MNC}  & ResNet-101-C4 & 24.6 & 44.3 & 24.8 & 4.7 & 25.9 & 43.6 \\
		\makecell[bl]{FCIS + OHEM}  & ResNet-101-dilated & 29.2 & 49.5 & - & 7.1 & 31.3 & 50.0 \\
		\hline
		\makecell[bl]{Mask R-CNN}  & ResNet-101-FPN & 36.9 & 59.0 & 39.5 & 19.9 & 39.7 & 48.3\\

		\makecell[bl]{MaskLab+ \cite{DBLP:journals/corr/abs-1712-04837}}  & ResNet-101 & 37.3 & 59.8 & 39.6 & 19.1 & 40.5 & 50.6\\
		
		\makecell[bl]{MaskLab+ \cite{DBLP:journals/corr/abs-1712-04837}}  & ResNet-101 (JFT) & 38.1 & 61.1 & 40.4 & 19.6 & 41.6 & 51.4\\
		
		\makecell[bl]{Mask Scoring R-CNN \cite{DBLP:journals/corr/abs-1903-00241}}  & ResNet-101-FPN & 38.3 & 58.8 &  41.5 & 17.8 &  40.4 & 54.4\\
		
		\makecell[bl]{HTC \cite{DBLP:journals/corr/abs-1901-07518}}  & ResNet-101-FPN & 39.7 & 61.8 & 43.1 & 21.0 & 42.2 & 53.5\\
		
		\makecell[bl]{\textbf{MaskPlus}}  & ResNet-101-FPN & 38.1 & 59.9 & 41.0 & 20.2 & 40.8 & 50.4 \\
		
		\makecell[bl]{\textbf{MaskPlus+}}  & ResNet-101-FPN & 40.9 & 63.0 & 44.5 & 23.5 & 43.6 & 52.3 \\
	\end{tabular}
	\vspace{4pt}
	\caption{\small{The overall performance of our MaskPlus framework, compared with previous methods (including newly-added concurrent works~\cite{DBLP:journals/corr/abs-1901-07518, DBLP:journals/corr/abs-1903-00241}) on the COCO test-dev subset. Final results are on the CodaLab COCO leaderboard. MaskPlus+ is achieved based on a simple cascade version of Mask R-CNN as described in~\cite{DBLP:journals/corr/abs-1901-07518, DBLP:journals/corr/abs-1712-00726}. Note that some other techniques such as ResNeXt-101, multi-GPU synchronized batch normalization, Atrous Spatial Pyramid Pooling, etc, that used in these concurrent works are not used in our work (thus there is potential for further improvement).}} 
	\label{final}
\end{table*}
\subsection{Overall Effectiveness of MaskPlus}

We adopted all methods in best settings introduced above to generate a final model, aka MaskPlus. We compare our MaskPlus with state-of-the-art approaches in the literature.
As shown in Table~\ref{final}, our approach clearly shows the state-of-the-art performance. 
To be specific, MaskPlus outperforms original Mask R-CNN in all provided metrics. And even compared with concurrent works \cite{DBLP:journals/corr/abs-1903-00241} \cite{DBLP:journals/corr/abs-1901-07518} which are also improved works based on Mask R-CNN, we can still give a competitive results. And we also create MaskPlus+ which is achieved based on a naive cascade version of Mask R-CNN as described in \cite{DBLP:journals/corr/abs-1712-00726} (also used in \cite{DBLP:journals/corr/abs-1901-07518}), which takes 0.5 times longer training steps to make model converge. Note that some other techniques such as ResNeXt-101, multi-GPU synchronized batch normalization,  Atrous Spatial Pyramid Pooling etc, that used in these concurrent works are not engaged in our work(thus there is potential for further improvement and collaboration.)

Finally, some results are visualized in Figure~\ref{example}. And our result can be seen on CodaLab COCO leaderboard.


\section{Conclusion} \label{sec:conclusion}
In this paper, we presented five methods for improving the mask generation in instance segmentation. It contains three novel  techniques  –  contextual  fusion, quasi-multitask learning and biased train, and we also extend the existing techniques, including boundary refinement, deconvolutional pyramid module, to further improve the accuracy. We incorporated these techniques into Mask R-CNN to build our MaskPlus framework, and conducted tests on the COCO dataset. The experiments demonstrate better results and shows the state-of-the-art performance. 




{\small
	\bibliographystyle{ieee}
	\bibliography{shichao_bib,shuyue_bib}
}

\end{document}